\pdfoutput=1
\documentclass[11pt]{article}

\usepackage[final]{acl}

\usepackage{times}
\usepackage{latexsym}
\usepackage{amsmath}
\usepackage{cleveref}
\usepackage{booktabs}
\usepackage{multirow}
\usepackage{pifont}
\usepackage{hyperref}       % hyperlinks
\usepackage{url}
\usepackage{amsfonts}
\usepackage{microtype}
\usepackage{graphicx}
\usepackage{colortbl}
\usepackage{amssymb}
\usepackage{tcolorbox}
\usepackage{tabularx}     % 用于创建宽度可控、内容可自动换行的表格
\usepackage{caption}      % 提供对图表标题的更多控制

\definecolor{Maroon}{rgb}{0.5, 0, 0}  % 自定义Maroon颜色
\definecolor{OliveGreen}{rgb}{0.33, 0.42, 0.18}  % 自定义OliveGreen颜色
\definecolor{Violet}{cmyk}{0.25, 0.5, 0, 0}
\usepackage[T1]{fontenc}
\usepackage[utf8]{inputenc}

\usepackage{microtype}

\usepackage{inconsolata}

\usepackage{graphicx}
\newcommand{\cmark}{\color{OliveGreen}{\ding{51}}}%
\newcommand{\xmark}{\color{Maroon}{\ding{55}}}%
\newcommand{\bcmark}{\color{Violet}{\ding{51}}}%
\newcommand{\notcheckmark}{\textcolor{black}{\bcmark\kern-1.1ex\raisebox{.7ex}{\rotatebox[origin=c]{125}{--}}}\color{black}}

\title{$\textsc{ChartEdit}$: How Far Are MLLMs From Automating Chart Analysis? Evaluating MLLMs' Capability via Chart Editing}

\author{
    \textbf{Xuanle Zhao\textsuperscript{1,}\footnotemark[1]},
    \textbf{Xuexin Liu\textsuperscript{1,}\footnotemark[1]},
    \textbf{Haoyue Yang\textsuperscript{1,}\footnotemark[1]},
    \textbf{Xianzhen Luo\textsuperscript{2}}, 
    \\
    \textbf{Fanhu Zeng\textsuperscript{1}},
    \textbf{Jianling Li\textsuperscript{3}}, 
    \textbf{Qi Shi\textsuperscript{4,}\footnotemark[2]}, 
    \textbf{Chi Chen\textsuperscript{4,}\footnotemark[2]}
    \\
\textbf{\textsuperscript{1}} Institute of Automation, Chinese Academy of Sciences, Beijing, China
\\
\textbf{\textsuperscript{2}} Harbin Institute of Technology, Harbin, China 
\\
\textbf{\textsuperscript{3}} Tianjin University, Tianjin, China
\textbf{\textsuperscript{4}} Tsinghua University, Beijing, China
\\
\texttt{\{zhaoxuanle2022, liuxuexin2022, yanghaoyue2024\}@ia.ac.cn}
}

\begin{document}
\maketitle

\renewcommand{\thefootnote}{\fnsymbol{footnote}}
% \footnotetext[2]{The two authors contribute equally to this work.}
\footnotetext[1]{Equal contribution.}
\footnotetext[2]{Corresponding authors.}
\renewcommand{\thefootnote}{\arabic{footnote}}

\begin{abstract}
Although multimodal large language models (MLLMs) show promise in generating chart rendering code, editing charts via code presents a greater challenge. This task demands MLLMs to integrate chart understanding and reasoning capacities, which are labor-intensive. While many MLLMs claim such editing capabilities, current evaluations rely on limited case studies, highlighting the urgent need for a comprehensive evaluation framework.
In this work, we propose \textsc{ChartEdit}, a novel benchmark designed for chart editing tasks, featuring $1405$ diverse editing instructions applied to $233$ real-world charts, each manually annotated and validated for accuracy. Utilizing \textsc{ChartEdit}, we evaluate the performance of 10 mainstream MLLMs across two types of experiments at both the code and chart levels.
The results suggest that large-scale models can generate code to produce images that partially match the reference images.
However, their ability to generate accurate edits according to the instructions remains limited. The state-of-the-art (SOTA) model achieves a score of only $59.96$, highlighting significant challenges in precise modification. In contrast, small-scale models, including chart-domain models, struggle both with following editing instructions and generating overall chart images, underscoring the need for further development in this area. Code is available at \url{https://github.com/xxlllz/ChartEdit}.
\end{abstract}

\section{Introduction}
\begin{figure}[t]
\centering
  \includegraphics[width=0.97\columnwidth]{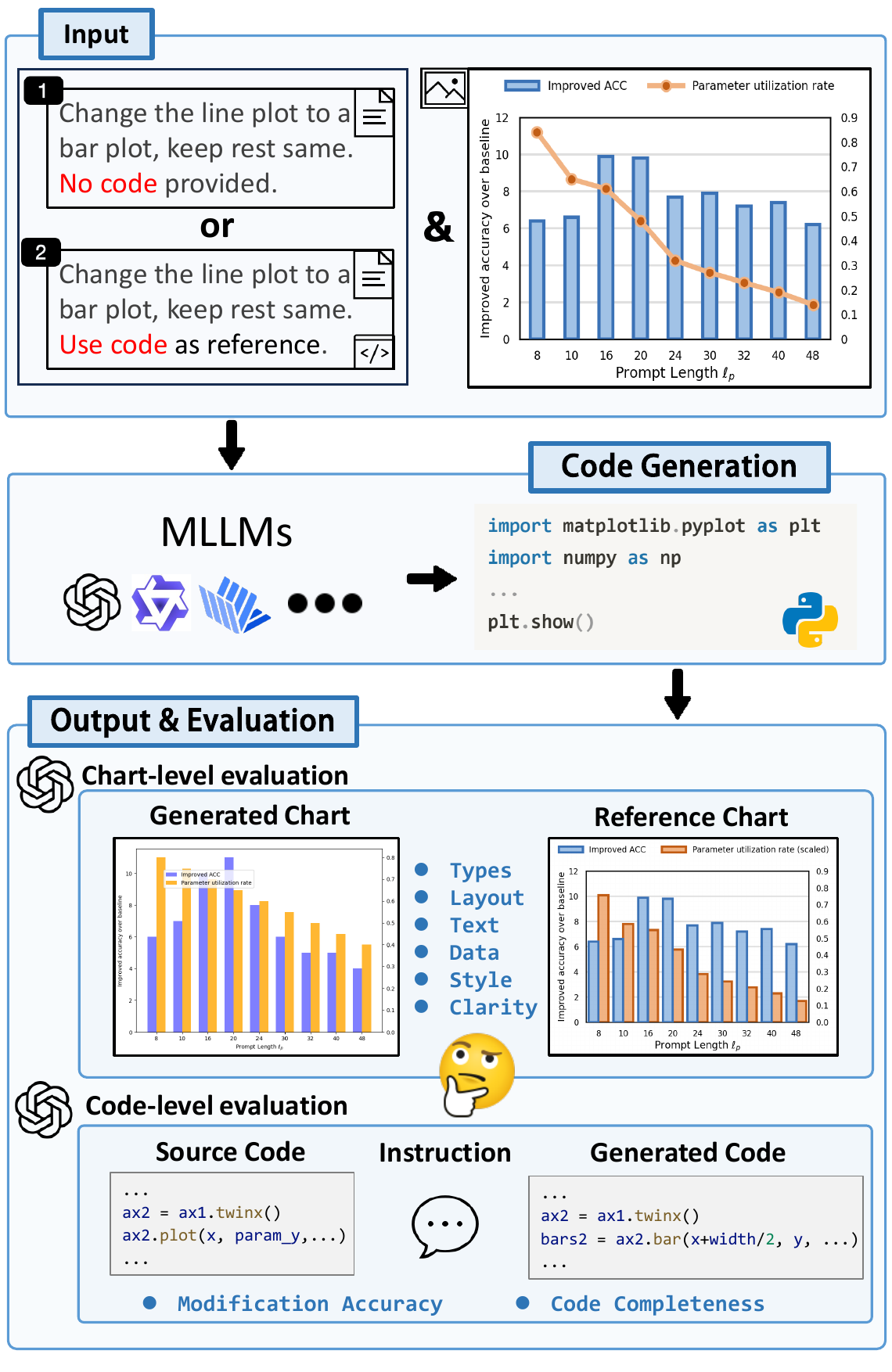}
  \caption{Overall pipeline. The inputs are a chart, editing instruction \textit{w/} or \textit{w/o} code. The MLLMs are instructed to generate the edited code. The final evaluation is constructed at both the code level and chart level.}
  \label{fig:overview}
  \vspace{-15pt}
\end{figure}

Data visualization is a crucial component in various fields, enabling individuals and organizations to present and interpret data effectively \cite{chen2008brief}. However, creating high-quality, visually appealing charts from scratch can be time-consuming and often requires extensive adjustments and references to documentation. This highlights the need for more efficient solutions to streamline the visualization process. Previous works have explored utilizing textual descriptions to generate visualization code automatically through the model \cite{yang2024matplotagent, zadeh2024text2chart31}, which significantly reduces the time and effort required for data visualization, making it more accessible and efficient for researchers.
% One potential approach to this process is to automatically generate chart code templates based on existing visualizations. This would allow researchers to customize them as needed instead of starting from scratch. 

Recently, multimodal large language models (MLLMs) \cite{liu2023improvedllava, zhang2024mm, guo2025hide, zhang2025branchlora}, which leverage the rich knowledge in large language models (LLMs) \cite{dubey2024llama}, have demonstrated impressive performance in processing and reasoning over various modalities, such as image \cite{wang2024visionllm, yu2025proglora}, video \cite{li2024llava} and speech \cite{wang2024cieasr}.
However, much of the existing research in code generation has primarily focused on using text as the sole input \cite{li2022competition, luo2023wizardcoder}, leaving the potential of multimodal information largely unexplored.

Although there have been recent efforts to explore multimodal code generation in specific areas like charts \cite{shi2024chartmimic, zhang2024gpt} and webpages \cite{yun2024web2code, si2024design2code}, these works typically address direct chart/web-to-code tasks, which are akin to using code for captioning. These approaches often lack diverse instructions and fail to account for interactive or iterative modifications. This raises an important question: Can current MLLMs function like expert chart analysts, generating modified chart code based on both the source chart and detailed editing instructions?

This is a far more complex challenge, as it requires MLLMs to not only extract relevant information from the chart but also generate corresponding code and adapt it according to the provided editing instructions. However, there is currently no high-quality, diverse benchmark for evaluating chart editing tasks. Previous works, such as ChartMimic \cite{shi2024chartmimic}, focus mainly on data-centric modifications, while the ChartLlama dataset lacks real-world charts, limiting the scope and relevance of their evaluations.

To address this gap, we introduce \textsc{ChartEdit}, a comprehensive evaluation benchmark consisting of $233$ real-world charts sourced from Arxiv, each accompanied by manually annotated code and $1405$ chart instructions with reference edited code.
The instructions and code are either human-written or initially generated by LLMs, followed by manual correction and alignment to ensure accuracy.
To ensure diversity, we pre-define $19$ chart types and six types of editing instructions. We also introduce two task formats, input chart \textit{w/} code or \textit{w/o} code, simulating real-world scenarios that require code editing from only the charts or charts with code.
Furthermore, we establish and evaluate metrics based on execution rate, code-level accuracy, and chart-level consistency to comprehensively assess MLLMs’ capabilities. These metrics measure the models’ ability to generate executable code, maintain editing precision, and ensure consistency across the generated visualizations.
We conduct experiments on three types of MLLMs: proprietary, general-domain open-source and chart-domain models. The results reveal that, despite notable advancements, state-of-the-art open-source models still show performance gaps compared to GPT-4o. Additionally, current small-scale MLLMs, including chart-domain models, continue to face significant challenges in generating editing code that accurately follows the instructions. Furthermore, we also investigate the impact of Chain-of-Thought (CoT) prompting \cite{wei2022chain} and analyze the models’ performance across various editing instructions and chart types.

\section{Related Works}

\subsection{Chart-Domain MLLMs}
Recently, MLLMs have achieved superior performance on many visual tasks by leveraging connectors to bridge the gap between large language models and vision encoders \cite{liu2023improvedllava, liu2023llava}. As a significant image type, chart-related tasks have received much attention. Previous works utilize a two-stage method that first extracts information from the chart and then utilizes language models to process the information \cite{liu2022deplot}. Currently, end-to-end MLLMs are utilized to solve chart-related tasks with a unified model. ChartLlama \cite{han2023chartllama} direct finetuning based on the existing LLaVA \cite{liu2023llava}. mPLUG-Owl \cite{Ye2023mPLUGOwlME} and mPLUG-Owl2 \cite{ye2024mplug} achieve superior performance on high-resolution chart images. ChartVLM \cite{xia2024chartx} utilizes a discriminator to determine whether intervention from the LLMs is required for a specific query. TinyChart \cite{zhang2024tinychart} proposes the token merging and PoT-based reasoning strategy to improve inference efficiency and understanding capacity.
ChartMoE \cite{xu2024chartmoe} utilize the Mixture-of-Expert (MoE) method to align many data formats with charts.

\begin{table*}[t]
\centering
\small
\setlength{\tabcolsep}{3pt}
\begin{tabular}{l|ccccccccccc}
\hline
\toprule
\multirow{2}{*}{\textbf{Name}} & \textbf{Output} & \textbf{w/ Real} & \textbf{w/ Cor.} &  \textbf{Diverse}  & \textbf{Open} & \multicolumn{5}{c}{\textbf{Editing Instruction}} \\
\cline{7-11}
 &  \textbf{Format} & \textbf{Chart} & \textbf{Code}  & \textbf{Types}  & \textbf{Domain} & \textbf{Data} & \textbf{Format} & \textbf{Layout} & \textbf{Style}  & \textbf{Text} \\
\midrule
ChartCraft \cite{yan2024chartreformer} & \texttt{Json} & \xmark & \xmark & \xmark & \xmark & \cmark & \cmark & \cmark & \cmark & \xmark  \\
Plot2Code \cite{wu2024plot2code}& \texttt{Code} & \cmark & \xmark & \cmark & \cmark & \xmark & \xmark & \xmark & \xmark & \xmark  \\
ChartX \cite{xia2024chartx}& \texttt{Code} & \xmark & \cmark & \cmark &  \cmark & \xmark & \xmark & \xmark & \xmark & \xmark  \\
AcademiaChart \cite{zhang2024gpt} & \texttt{Code} & \cmark &  \xmark & \xmark & \cmark & \xmark & \xmark & \xmark & \xmark & \xmark \\
ChartMimic \cite{shi2024chartmimic} & \texttt{Code} & \cmark & \cmark & \cmark & \cmark & \cmark &  \xmark & \xmark & \xmark  & \xmark \\
\midrule
\textsc{ChartEdit (ours)} & \texttt{Code} & \cmark & \cmark & \cmark & \cmark & \cmark & \cmark & \cmark & \cmark & \cmark  \\
\bottomrule
\end{tabular}
\vspace{-5pt}
\caption{The comparison of our proposed \textsc{ChartEdit} evaluation benchmark with other chart-related benchmarks. \textsc{ChartEdit} is the first evaluation benchmark, and it contains various editing instructions and reference codes.}
\label{tab:comparison_benchmarks}
\vspace{-5pt}
\end{table*}

\subsection{MLLMs For Code}
Previous advancements in LLMs have significantly contributed to code-related tasks. Notable models, such as DeepSeek Coder \cite{guo2024deepseek} and StarCoder \cite{lozhkov2024starcoder}, have demonstrated substantial progress in tasks like code generation and error fixing. However, these models typically operate in a single-modal setting, relying on textual input, which limits their ability to address the complexity and range of problems. 

Multimodal code generation has recently gained significant attention.
Several works, like Design2Code \cite{si2024design2code} and Web2Code \cite{yun2024web2code}, focus on assessing the capacities of MLLMs in generating HTML code for web pages. RoboCodeX \cite{mu2024robocodex} proposes a multimodal code generation framework for robot behaviour synthesis. The field of chart-to-code generation has also attracted considerable attention \cite{zhao2025chartcoder}, as it focuses on generating code that accurately reproduces a given chart image. This task is challenging due to the complex visual elements in charts, demanding advanced models to accurately translate them into functional code. Recent works \cite{wu2024plot2code, xia2024chartx, zhang2024gpt} evaluate the capacities of MLLMs in this context. 
% However, these benchmarks generally focus on direct generation tasks, lacking further evaluation of the capacities of MLLMs under diverse instructions, such as chart modifications.
Although some works like \cite{xia2024chartx, shi2024chartmimic} have considered this problem, the editing instructions in their works only focus on one type of modification, which is not diverse enough.

\begin{figure*}[t]
\label{fig:main}
  \includegraphics[width=0.99\linewidth]{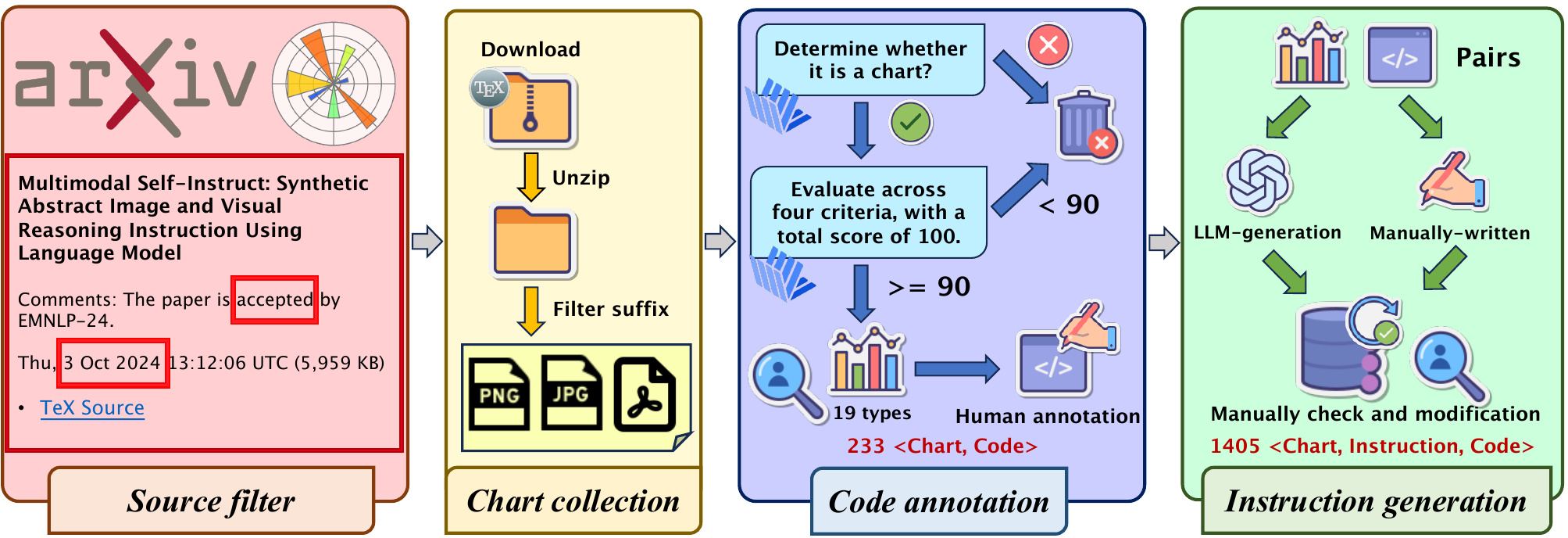}
  \caption {The pipeline for constructing the \textsc{ChartEdit} evaluation dataset begins with filtering and crawling ArXiv papers based on keywords found in their comments. After that, we remove irrelevant files by filtering out specific suffixes. We then use an MLLM to distinguish and filter out non-chart images, scoring the remaining images. These images are further screened based on these scores and reviewed by human evaluators. Also, Code annotations are manually written by human evaluators. The editing instructions and reference edited code are constructed utilizing two strategies: one based on LLM and the other manually written. Finally, all the <Chart, Instruction, Code> triplets are reviewed and modified by human evaluators to enhance correspondence and accuracy.
  }
  \vspace{-5pt}
\end{figure*}

\section{\textsc{ChartEdit}}
In this section, we first introduce the definition of the chart editing task and illustrate the data collection and organization pipeline of \textsc{ChartEdit}.
\subsection{Task Definition}
In this work, we aim to leverage MLLMs to edit charts and generate the corresponding code as instructed. The task can be summarized as follows: given a chart image $X$ from Arxiv papers, along with an editing instruction $I$, the model is expected to generate the corresponding visualization code after applying the edits, regardless of whether the original code $C$ is provided. Thus, the task can be represented as:
\begin{equation}
    O = M(X\vee(X, C), I)
\end{equation}
where $M$ denotes the processing model. $X\vee(X, C)$ indicates that the model can either receive $X$ (Chart \textit{w/o} Code) or $(X, C)$ (Chart \textit{w/} Code) as input. $O$ is the output code in \texttt{Python}, which utilize libraries like \text{Matplotlib} and \texttt{Seaborn} to plot the edited chart.

\subsection{Data Construction}
\subsubsection{Chart Collection and Filtering}
In this study, to collect real-world chart images, we begin by crawling papers published on Arxiv using web scraping tools like BeautifulSoup.
To ensure the quality of the collected images, we first retrieve the IDs and comments of papers published on Arxiv, then filter the papers by selecting those whose comments include keywords such as “submit”, “accept”, “under review” and “camera ready”. After this filtering step, we use the ArXiv API to download the LaTeX source files for the selected papers.
After downloading the source files, we remove irrelevant files based on their suffixes, retaining only those ending in \texttt{.png}, \texttt{.pdf}, \texttt{.jpg}, and \texttt{.svg}. However, many of these images are not charts that can be reproduced by visualization code. To filter out high-quality chart images, we employ a two-step process. First, we use an MLLM to assess whether each image is a chart through zero-shot prompting. Despite this assessment, we found that many low-quality chart images remained. To address this, we developed a scoring mechanism, instructing the MLLM to evaluate the images based on four criteria: \textit{Aesthetics}, \textit{Readability}, \textit{Reproducibility}, and \textit{Data Presentation Simplicity}. Each image is assigned a total score of $100$, and we filter out those scoring below $90$. After completing this two-step filtering process, we are left with more than $1,000$ high-quality chart images.

\subsubsection{Code Annotation}
While we have collected a sufficient number of chart images, the charts crawled from Arxiv do not include the code required for reproduction. To refine our dataset and facilitate the extraction of edited code, we manually filter and write Python code that can reproduce the corresponding charts. However, some chart types are still missing. To address this, we supplement the dataset with additional charts from other platforms, such as Kaggle and the Matplotlib gallery \cite{hunter2007matplotlib}. To prevent data leakage, we manually modify the code to generate visually distinct versions of the original charts. As a result, we have successfully obtained $233$ charts, along with their corresponding code, forming <chart, code> source pairs.

\subsubsection{Instruction Generation}\label{Instruction generation}
After obtaining the chart source code, we propose two approaches for constructing editing instructions: (1) LLM-based generation and (2) human-written instructions. In the LLM-based approach, we predefine five key editing categories: \textit{style}, \textit{format}, \textit{layout}, \textit{data}, and \textit{text}, each with multiple specific subtypes. By providing the LLM with both the source code and the chart image, we prompt it to generate editing instructions and the corresponding edited code. To ensure a diverse set of outputs and minimize errors, we generate at least three variations for each chart and instruction type. To ensure the instructions and code are rigorous and aligned with human expectations, we manually verify and modify all instructions, code, and edited images for consistency. This process results in a total of $1,172$ <chart, instruction, code> triplets. While LLM-based generation can produce a large volume of triplets, the diversity of descriptions and their relevance to real-world needs may be constrained by inherent biases in LLM outputs. To mitigate this limitation, we manually write one editing instruction for each chart with reference code, ensuring that the dataset more closely aligns with real-world requirements.

\begin{figure}[t]
\centering
  \includegraphics[width=0.85\columnwidth]{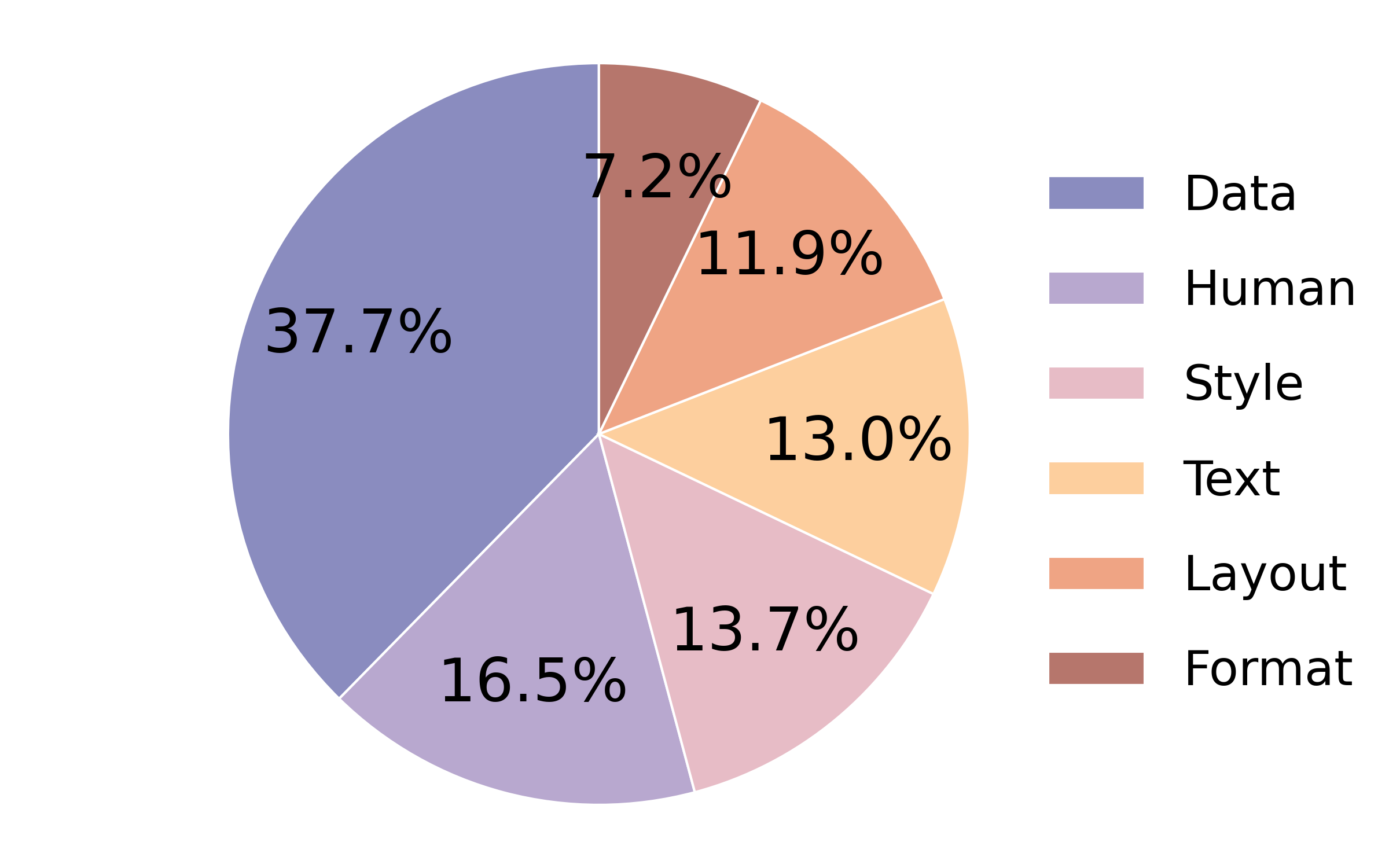}
  \caption{The number and specific proportions of different types of editing instructions in our \textsc{ChartEdit} evaluation benchmark.}
  \label{fig:pie_distribution}
  \vspace{-10pt}
\end{figure}

\subsection{Dataset Statistics and Analysis}
\begin{figure}[t]
  \includegraphics[width=\columnwidth]{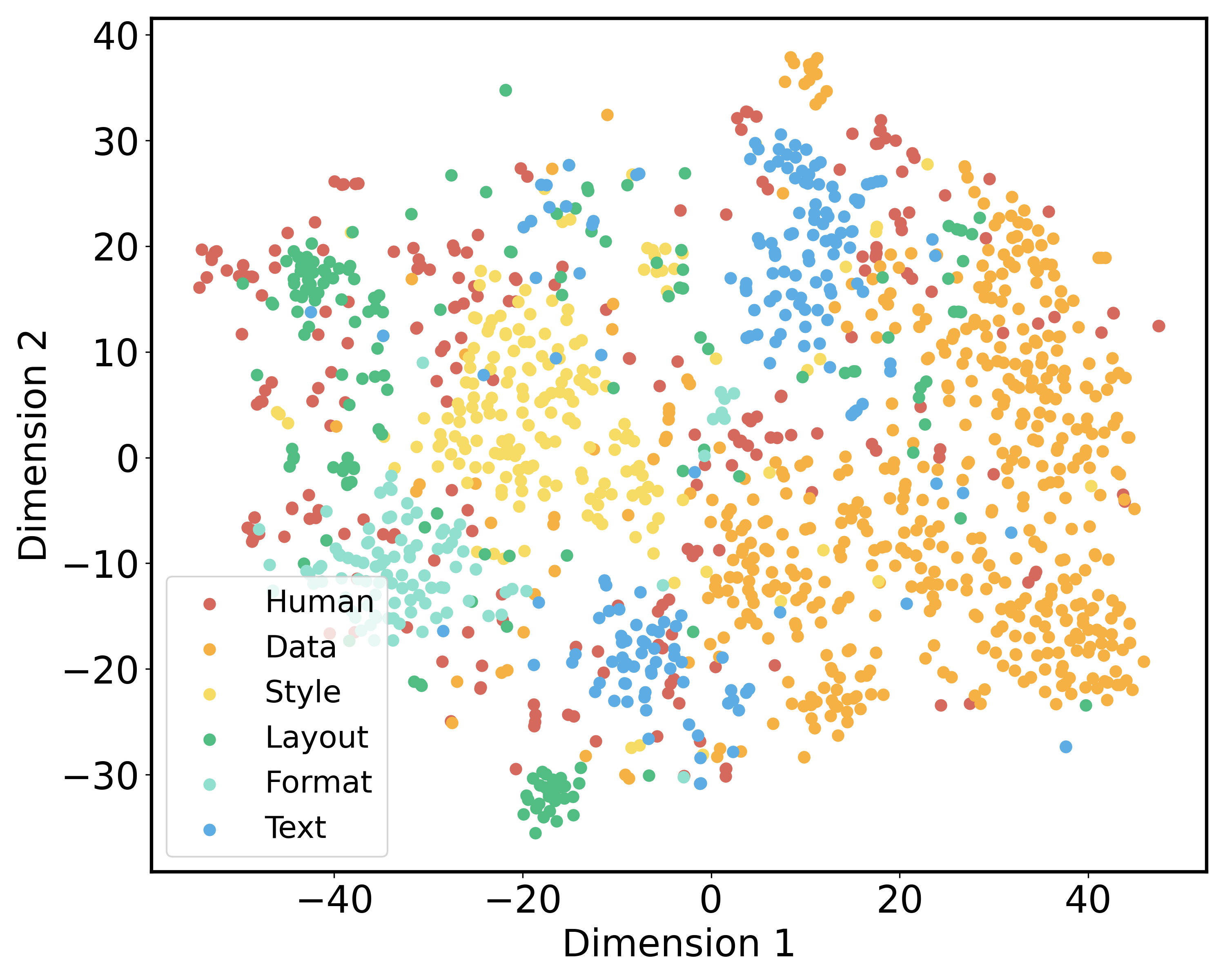}
  \caption{The dimension reduction of the editing instructions in \textsc{ChartEdit} with various colors represents different editing types. We choose the Sentence-BERT \cite{reimers2019sentence} as the embedding model.}
  \label{fig:dimension_redcution}
  \vspace{-10pt}
\end{figure}
Using LLM generation and manual annotation, we construct our evaluation benchmark, \textsc{ChartEdit}, which comprises $233$ <chart, code> pairs and $1,405$ <chart, instruction, code> triplets. We further analyze the diversity of \textsc{ChartEdit} from both chart and instruction perspectives.
To ensure a diverse selection of chart types, we have predefined $19$ distinct chart categories, carefully curating instances for each type. The distribution of these chart types is provided in the \Cref{chart_statictic}. Additionally, we enhance the diversity of editing instructions by classifying them into six categories: five generated by LLMs, as outlined in \Cref{Instruction generation}, and one written by humans.
The proportions of each editing category are shown in \Cref{fig:pie_distribution}. In addition to exploring the diversity of editing types, we also examine the different formats in which these instructions are presented, including plain text descriptions, code symbols such as \texttt{marker='*'}, and hexadecimal color codes like \texttt{color=’\#ABCDEF’}. We embed these editing instructions and compare them with instructions from other chart-to-code benchmarks using dimensionality reduction techniques. The result in \Cref{fig:dimension_redcution} shows embedding after dimension reduction, which indicates that different editing types are generally separated, and the human-written instructions almost cover all the instruction types.
% Unlike direct chart-to-code generation, the chart editing task requires models to have a deeper understanding of the instructions and chart before generating the corresponding code.

\section{Experiments}
\subsection{Baselines}
To facilitate a more effective comparison of existing MLLMs in chart editing tasks, we benchmark three categories of widely used MLLMs: proprietary, open-source general-domain, and chart-domain models.
\textbf{(1) Proprietary Models}: We evaluate two top-performing models in the multimodal domain: GPT-4o \cite{openai2024gpt4o} and Claude-3.5-Sonnet \cite{anthropic2024claude} which represent the cutting edge in the multimodal domain.
\textbf{(2)	Open-Source General-Domain Models}: We select five competitive open-source models from a variety of sizes, listed in descending order of total parameters. These models include InternVL-V2.5-78B \cite{chen2024expanding}, Qwen2-VL-72B \cite{wang2024qwen2}, LLaVA \cite{liu2023llava}, InternVL-V2.5-8B and Phi-3.5-Vision \cite{abdin2024phi}. These models represent the base models commonly used in the multimodal domain.
\textbf{(3) Chart-Domain Models}: For this category, we focus on models already designed for chart-to-code generation tasks, including ChartLlama \cite{han2023chartllama}, TinyChart \cite{zhang2024tinychart}, and ChartMoE \cite{xu2024chartmoe}. 

For the evaluation, we use the direct instruction prompting method across all models. Detailed prompts and implementation steps are provided in the \Cref{fig:code_generation_prompt}.

\begin{table}[t]
  \centering
  \begin{tabular}{lc}
    \hline
    \textbf{Statistic} & \textbf{Value} \\
    \hline
    \textbf{Charts} \\
    Total Chart      & 233           \\
    Types            & 19 \\
    Average size (px) & 876$\times$575 \\
    Maximum size (px) & 3000$\times$1600 \\
    \hline
    \textbf{Instruction} \\
    Total Instruction      & 1405           \\
    Edit Types     & 6           \\
    \hline
    \textbf{Code} \\
    Average Length & 650 \\
    Maximum Length & 3310 \\
    \hline
  \end{tabular}
  \vspace{-5pt}
  \caption{ChartEdit dataset statistics. The code length is calculated based on the number of tokens utilizing the Llama3.2 tokenizer \cite{dubey2024llama}.}
  \label{tab:accents}
  \vspace{-5pt}
\end{table}

\subsection{Evaluation Metrics}
Given the lack of established evaluation metrics for assessing the completeness of image editing, we draw inspiration from recent research that uses LLMs as evaluators \cite{gu2024survey, zheng2024judging}. In this work, we leverage \texttt{GPT-4o} to evaluate two key aspects: first, \textbf{\textit{whether the generated code aligns with the provided editing instructions}}, and second, \textbf{\textit{whether the generated code effectively produces the intended chart}}.

\begin{table*}[t]
\small
\setlength{\tabcolsep}{4pt}
\centering
\begin{tabular}{l|ccc|ccc}
\toprule
\multirow{3}{*}{\textbf{Model}}  & \multicolumn{3}{c|}{\textbf{Chart \textit{w/o} Code}} & \multicolumn{3}{c}{\textbf{Chart \textit{w/} Code}} \\
\cmidrule{2-7} 
& \multicolumn{1}{c}{Exec.Rate} & \multicolumn{1}{c}{Code-Level} & Chart-Level & Exec.Rate & \multicolumn{1}{c}{Code-Level} & \multicolumn{1}{c}{Chart-Level}\\ 
\midrule
\multicolumn{7}{c}{\it{Proprietary Models}} \\ 
GPT-4o \cite{openai2024gpt4o} & \textbf{91.46} & \textbf{59.96} & \textbf{79.87} & \textbf{98.89} & 89.96 & \textbf{93.68} \\
Claude-3.5-Sonnet \cite{anthropic2024claude} &88.22 &47.32& 54.92 & 89.50& 88.99&81.68\\ 
\midrule
\multicolumn{7}{c}{\it{Open-Source General-Domain  Models}} \\
InternVL2.5-78B \cite{chen2024expanding} & 79.66 & 55.67 & 70.77 & 94.31 & \textbf{92.56} & 92.81  \\
Qwen2-VL-72B \cite{wang2024qwen2} & 81.65 & 46.67 & 64.06 & 86.09 &90.30&79.51\\
InternVL2.5-8B \cite{chen2024expanding} & 62.37 & 39.24 & 45.85 &85.20 & 90.16 & 87.36 \\
Phi3.5-Vision \cite{abdin2024phi} & 67.13 &40.19 & 37.65 & 74.44& 82.31&89.16\\
LLaVA-13B \cite{liu2023llava} & 48.75 & 11.35 & 16.71 & 30.44&71.88 & 25.82\\
\midrule
\multicolumn{7}{c}{\it{Chart-Domain  Models}} \\
ChartMoE \cite{xu2024chartmoe} &53.44  & 21.60 & 34.73 & 81.89 & 84.87 & 89.31\\
ChartLlama \cite{han2023chartllama} & 52.30 & 15.82& 27.42 & 46.34 &50.18& 47.00 \\
TinyChart \cite{zhang2024tinychart}& 36.13 & 18.34 & 25.09 & 3.51 &18.40&2.54 \\
\bottomrule 
\end{tabular}
\vspace{-5pt}
\caption{Evaluation results of various baseline models on Chart \textit{w/o} Code and Chart \textit{w/} Code tasks. The performance is evaluated from three aspects: the code Execution Rate, Code-Level, and Chart-Level scores. The best performances are indicated in \textbf{bold}.}
\vspace{-5pt}
\label{tab:main_results_direct}
\end{table*}
Therefore, we propose evaluating both the code level and the chart level. The code-level evaluation is based on the source code, editing instructions, and output code. We use \texttt{GPT-4o} to score the models on two aspects: \textit{Modification Accuracy} and \textit{Code Completeness}.
Building on related approaches \cite{shi2024chartmimic, zhang2024gpt}, we also evaluate the model’s overall generation capabilities at the chart level. This evaluation is performed by comparing how closely the generated chart reproduces the reference edited chart (manually created). The results reflect the degree of completeness of the generated chart image. Following \cite{shi2024chartmimic}, we also utilize \texttt{GPT-4o} to evaluate six aspects: \textit{Types}, \textit{Layout}, \textit{Text}, \textit{Data}, \textit{Style}, and \textit{Clarity}. Detailed information is provided in the \Cref{fig:chart_level_prompt}.

\begin{table}[t]
\setlength{\tabcolsep}{3pt}
\small
\centering
\begin{tabular}{l|cccccc}
\toprule
Model & Types & Layout & Text & Data & Style & Clarity\\
\midrule
% Full score & 20 &10 &20 &20 &20 & 10 \\
GPT-4o& 18.35 & 9.49 & 15.26  & 14.87 & 13.55 & 8.53\\
% \midrule
\scalebox{0.8}{InternVL2.5-78B} & 16.25 & 9.03 & 13.13  & 13.59 & 11.12 & 7.91\\
\scalebox{0.85}{InternVL2.5-8B} & 10.22 & 7.68 & 8.21  & 7.21 & 6.19 & 6.34\\
TinyChart & 4.99 & 4.60 & 4.39 & 3.71 & 3.41 & 4.16 \\
\bottomrule 
\end{tabular}
\vspace{-5pt}
\caption{Detailed results of Chart-Level scores on Chart \textit{w/o} Code task.}
\label{tab:detail_results_chart}
\vspace{-5pt}
\end{table}

\begin{table}[t]
\setlength{\tabcolsep}{3pt}
\small
\centering
\begin{tabular}{l|cc}
\toprule
Model & Modification Acc  & Code Complete \\
\midrule
% Full score & 20 &10 &20 &20 &20 & 10 \\
GPT-4o & 30.67 & 29.29 \\
% \midrule
InternVL2.5-78B & 29.05 & 26.62 \\
InternVL2.5-8B & 20.16 & 19.08 \\
TinyChart & 6.46 & 11.88 \\
\bottomrule 
\end{tabular}
\vspace{-5pt}
\caption{Detailed results of Code-Level scores on Chart \textit{w/o} Code task.}
\label{tab:detail_results_code}
\vspace{-5pt}
\end{table}

\subsection{Main Results}
The main results for all the MLLMs are presented in \Cref{tab:main_results_direct}. Although open-source models have made significant strides, and in some cases even outperformed GPT-4 in various tasks \cite{masry2022chartqa, zhang2024mm}, there is still a noticeable gap when it comes to handling complex multimodal tasks \cite{wang2024charxiv, shi2024chartmimic}. In our experiments, GPT-4o delivers the best performance on the Chart \textit{w/} Code task, achieving the highest Execution Rate, Code-Level, and Chart-Level scores. However, the Code-Level score is still not high enough, indicating that GPT-4o faces challenges with precise editing. Among open-source MLLMs, InternVL2.5-78B achieves the best performance, but it still lags behind GPT in terms of code execution rate and chart-level metrics. Our analysis shows that the largest gap between proprietary and open-source models lies in generating the complete code corresponding to an image. There is little difference in instruction-following ability between the two, and both struggle with detailed modifications in chart editing. The chart-domain models perform worse in this task, and we believe this is due to the current limitations of MLLMs in the chart-domain, particularly in handling editing tasks. Although TinyChart \cite{zhang2024tinychart} and ChartLlama \cite{han2023chartllama} have been fine-tuned on relevant datasets, both lack the ability to follow editing instructions effectively. The instructions in chart editing are much more diverse than those in direct chart-to-code generation tasks.

In the Chart \textit{w/} Code task, providing the source code of the input chart leads to significant performance improvements across all proprietary and open-source general-domain models. These models can generate more accurate code, which we believe is due to the source code supplying essential information, such as the chart’s data, text, and style. This significantly alleviates the challenge for the model in extracting relevant information from the chart itself. 
In this context, the task becomes more like editing the code with the chart images serving as auxiliary context, which is much easier than the Chart \textit{w/o} Code task. 
However, we observe that most Chart-Domain models perform significantly worse in this setting. Upon analyzing TinyChart, we found that it lacks training data for relevant tasks, especially those involving longer code snippets in the input. ChartLlama claims to support chart editing, but the diversity of chart types and editing instructions in its training data is limited. As a result, its performance decreases slightly when faced with inputs that include code. 
Overall, tasks with code are generally easier than those without, denoting chart-domain models still require substantial improvements in handling diverse instructions.

The detailed scores of the chart-level are presented in \Cref{tab:detail_results_chart}, and we select four models for analysis. Since the Chart \textit{w/o} Code task is significantly more challenging, we only provide detailed scores for this task. The result shows that the performance gaps between proprietary and open-domain models are most noticeable in the text, data, and style aspects, which are key pieces of information in the chart. \Cref{tab:detail_results_code} presents the detailed scores at the code level, highlighting that there remains a gap between the state-of-the-art open-source models and proprietary models in generating complete code. At the same time, the large-scale model has a significant advantage over the small-scale model in terms of modification accuracy.

Furthermore, we also recruit human evaluators to score the editing results of four popular MLLMs. Details about the results are listed in \Cref{human_evaluation}.
\section{Discussion}

\begin{figure}[t]
  \includegraphics[width=\columnwidth]{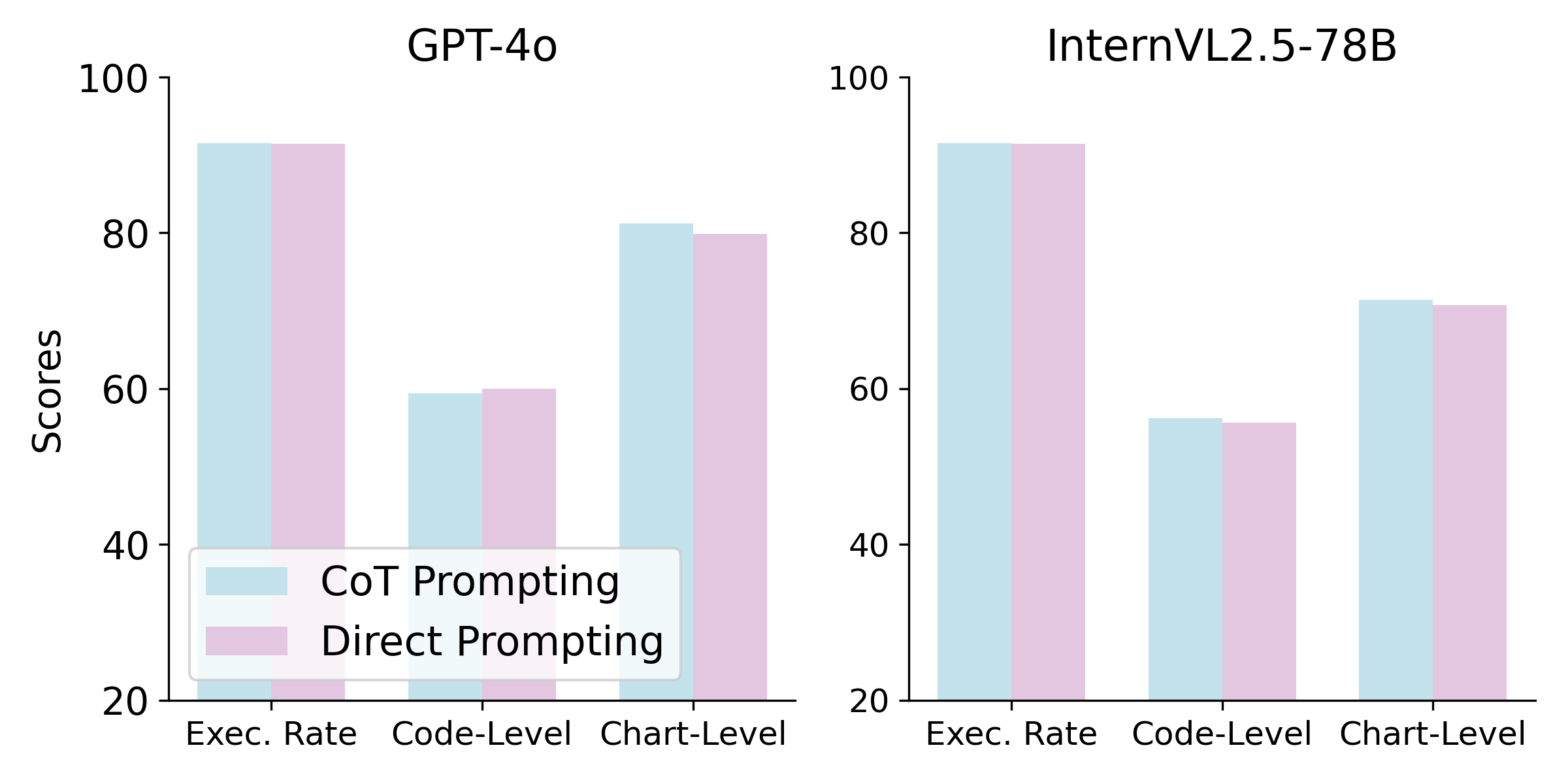}
  \caption{The result comparisons between GPT-4o and InternVL2.5-78B in direct and zero-shot Cot prompting setting. In most cases, the effect of CoT prompting shows a negligible improvement over direct prompting.}
  \label{fig:analysis_cot}
  \vspace{-10pt}
\end{figure}
In this section, we conduct many analyses to answer the following questions. 

\textbf{RQ1: Is Chain-of-Thought (CoT) prompting useful for MLLMs in chart editing tasks for proprietary models?} 
Answer: \textbf{No}. In the CoT setting, we provide MLLMs with zero-shot prompting, instructing them to first understand the source chart, then analyze the editing instructions, and finally generate the code. However, the result in \Cref{fig:analysis_cot} shows whether using proprietary models or open-source models, CoT prompting shows almost no improvement in performance. We analyze the thought process of GPT-4o and find that while CoT helps the model better understand the components in the chart and the editing instructions, it does not result in more accurate code generation. The detailed CoT prompts are provided in the \Cref{fig:code_generation_prompt}.

\begin{figure}[t]
  \includegraphics[width=\columnwidth]{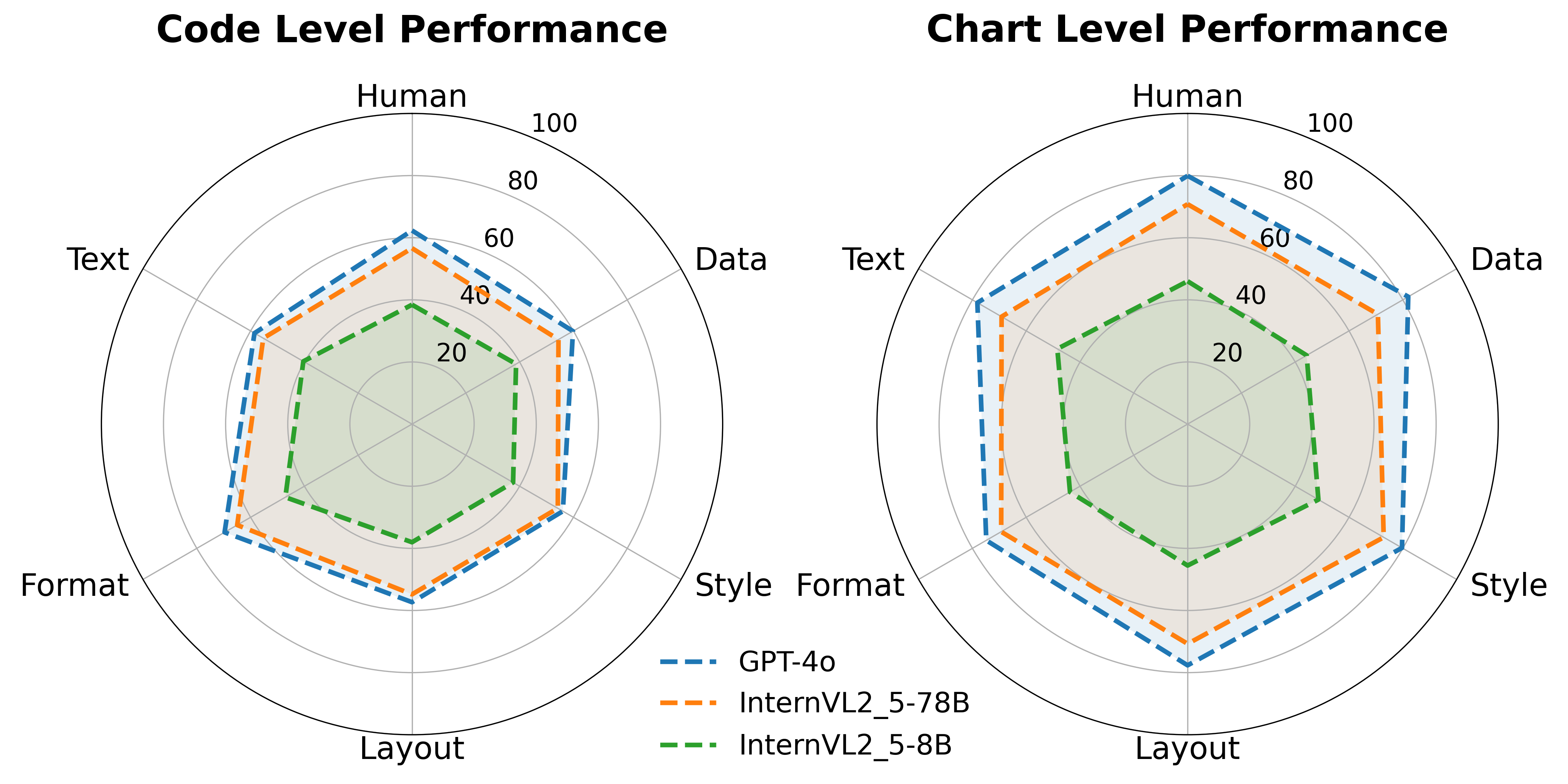}
  \caption{A comparison of task performance among GPT-4o, InternVL2.5-78B, and InternVL2.5-8B across both code-level and chart-level tasks in various instruction categories.}
  \label{fig:analysis_editing_type}
\end{figure}

\begin{figure}[t]
  \includegraphics[width=\columnwidth]{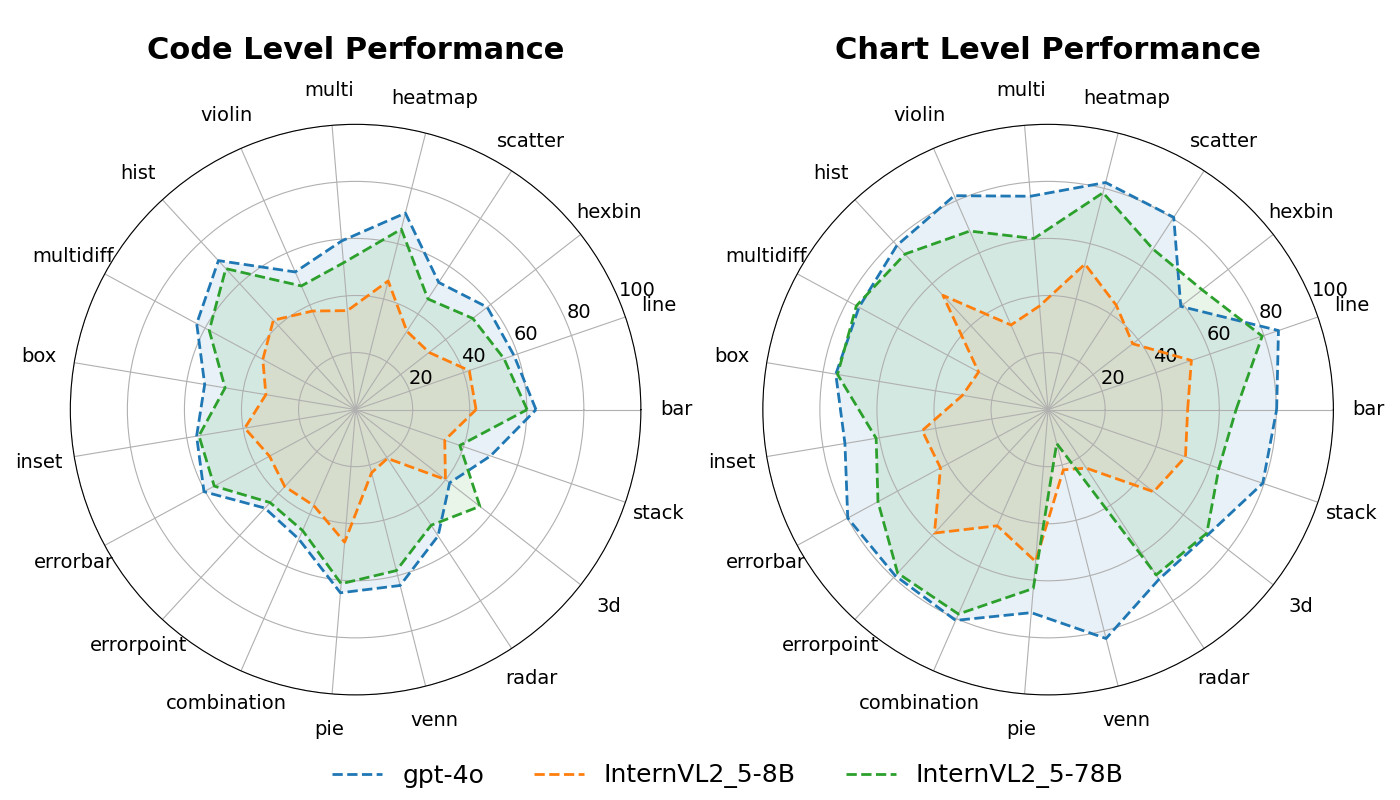}
  \caption{A comparison of task performance among GPT-4o, InternVL2.5-78B, and InternVL2.5-8B across both code-level and chart-level tasks in various chart categories.}
  \label{fig:analysis_chart_type}
  \vspace{-10pt}
\end{figure}

\begin{figure*}[t]
  \includegraphics[width=0.99\linewidth]{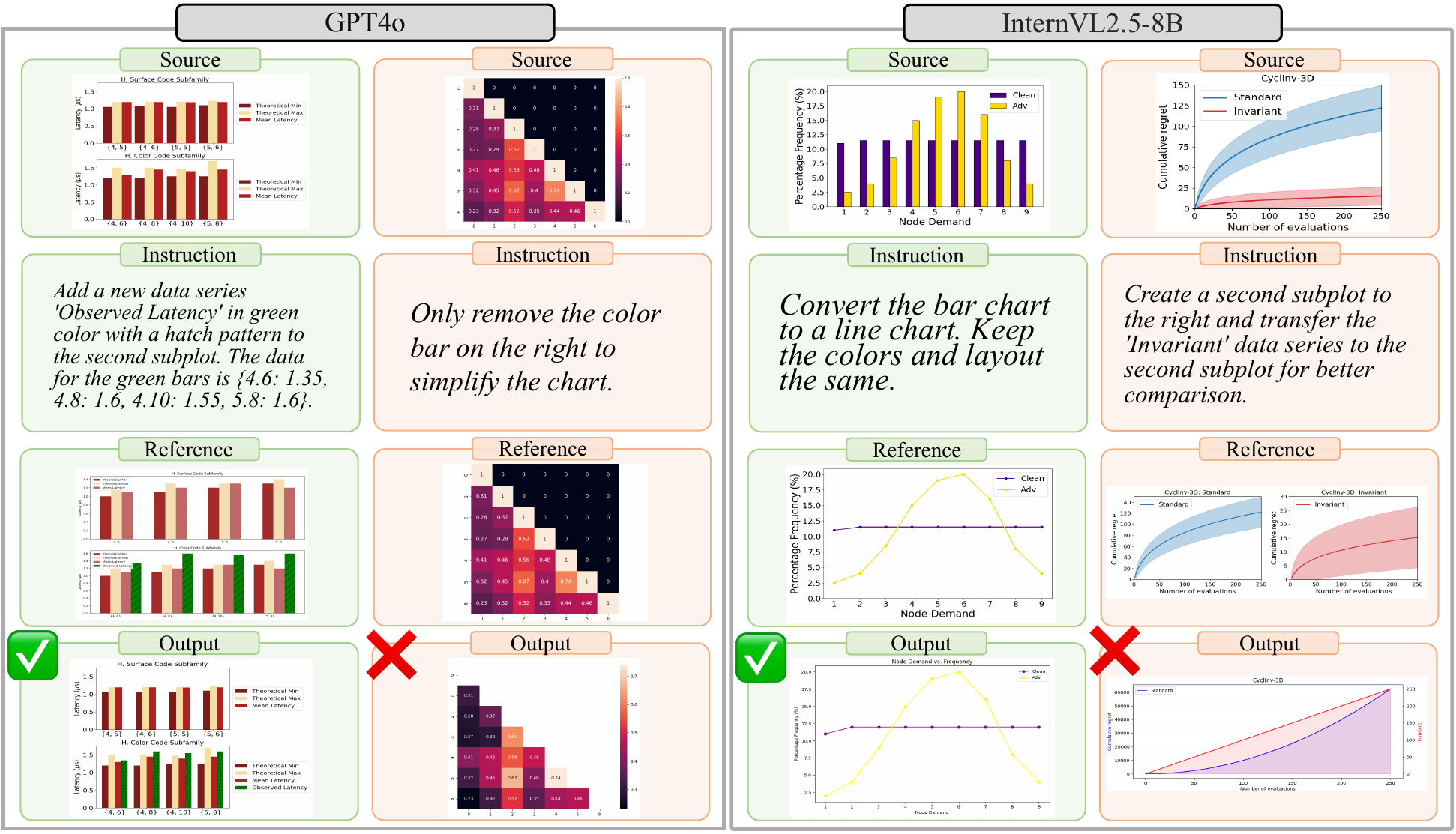}
  \caption {Case studies comparing the state-of-the-art proprietary model GPT-4o with a popular used open-source general-domain model InternVL2.5-8B. Generally, GPT-4o could perform better in a wide range of chart editing tasks, although there are still some tasks it struggles with. InternVL2.5-8B performs reasonably well in handling common chart types like bar and line charts, but it struggles with more complex editing instructions.}
  \vspace{-10pt}
\end{figure*}

\textbf{RQ2: How does the performance of MLLMs vary across different types of editing instructions?} In \Cref{fig:analysis_editing_type}, we compare the performance across different types of editing tasks and find that models are much more effective at generating code to modify chart types. Our analysis reveals that unlike tasks requiring first grounding and then modification of specific elements, the format conversion task focuses on altering the overall visual representation of the chart, which makes it easier to edit the code. However, at the chart level, model performance on format conversion tasks is limited. This suggests that while models can change the chart type effectively, they still struggle with more complex modifications involving other elements and fail to capture all the details of the chart.

\begin{figure}[t]
  \includegraphics[width=\columnwidth]{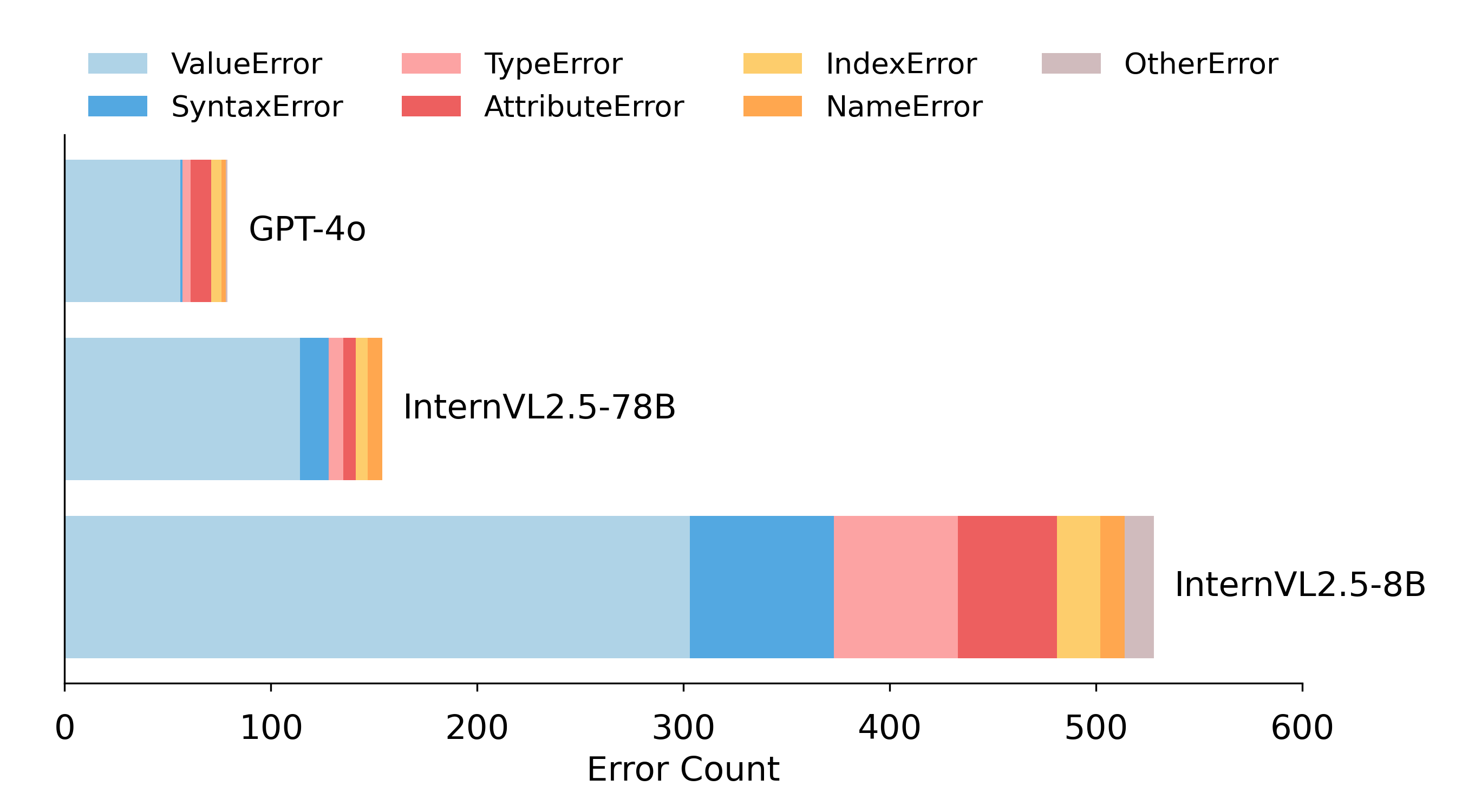}
  \caption{A comparison of error types of output code generated by GPT-4o, InternVL2.5-78B, and InternVL2.5-8B.}
  \label{fig:analysis_error_type}
  \vspace{-10pt}
\end{figure}

\textbf{RQ3: How does the performance of MLLMs vary across different types of charts?}
\Cref{fig:analysis_chart_type} illustrates model performance across different chart types. The results show that the most significant performance discrepancy between open-source and proprietary models occurs with \texttt{venn} diagrams. Additionally, noticeable differences emerge in handling more complex chart types, such as \texttt{errorbar} and \texttt{multi} charts. Upon further analysis, we found that the comparable performance of InternVL2.5-7B and GPT-4o on the code-level tasks is since this metric primarily evaluates the models’ capacity to follow instructions. Since most editing instructions for \texttt{venn} diagrams do not require altering the chart type itself, the performance gap at the code level is smaller than at the chart level.

\textbf{RQ4: How do the error types of generated code vary across MLLMs?}
\Cref{fig:analysis_error_type} shows the types of runtime errors generated by different models. The results reveal that \texttt{ValueError} constitutes the largest proportion of errors, accounting for more than half of the total. Interestingly, the proportion of errors excluding \texttt{ValueError} tends to decrease as the model’s performance improves. Our analysis suggests that as code generation capabilities improve, many common errors are effectively mitigated while more complex issues related to data processing persist.

\section{Conclusion}
In this work, we introduce \textsc{ChartEdit}, a high-quality benchmark that includes various types of editing instructions, each of which is either manually written or first generated by an LLM and then manually corrected. We evaluate performance across proprietary, open-source general-domain, and chart-domain models. The results demonstrate that proprietary models consistently outperform others in both following editing instructions and generating accurate chart images. Additionally, we observe that current chart-domain models generally struggle with complex instructions and handling diverse inputs. We believe that automated chart editing in academic research holds great promise, but MLLMs need to further improve their ability to effectively process chart images.

\section*{Limitation}
From our perspective, our work has several limitations: (1) Our works only consider textual prompts such as Chain-of-Thought. However, more visual prompt methods could be evaluated. (2) The dataset size may not be enough in some situations. Maybe a much larger evaluation benchmark could help to find more interesting findings. (3) More accurate evaluation methods. Our code-level and chart-level metrics are evaluated via LLM, so a better calculation method could be proposed.

\section*{Ethical Statement}
Our research employs publicly available models and data with proper citations. This approach minimizes the risk of generating toxic content, leveraging the widely used and non-toxic nature of our datasets and prompts.

\bibliography{custom}

\clearpage
\appendix

\section{Appendix}
\label{sec:appendix}
\subsection{Instruction Generation Details}
In data generation, we use InternVL2.5-78B to evaluate whether the images crawled from Arxiv are charts and to score the quality of these chart images. Charts with scores above 90 pass the initial filtering.

\subsection{Chart Statictic}\label{chart_statictic}
We present the count of each chart type in \Cref{table:chart_type_counts}. The largest proportions are observed in bar and line charts, which are the most common chart types in Arxiv papers.

\begin{table}[h]
\centering
% \small
\begin{tabular}{cc|cc}
\toprule
\textbf{Chart Type} & \textbf{Count} & \textbf{Chart Type} & \textbf{Count} \\
\midrule
line & 62 & bar & 41 \\
multi & 22 & heatmap & 21 \\
radar & 13 & scatter & 12 \\
errorbar & 10 & pie & 10 \\
violin & 7 & combination & 6 \\
box & 5 & errorpoint & 4 \\
hist & 4 & 3d & 3 \\
inset & 3 & multidiff & 3 \\
hexbin & 2 & stack & 2 \\
venn & 2 & & \\
\bottomrule
\end{tabular}
\caption{\centering Count of Each Chart Type in the Dataset} % 添加 \centering 使 caption 居中
\label{table:chart_type_counts}
\vspace{-10pt}
\end{table}

\subsection{Human Evaluation}\label{human_evaluation}
We recruit human evaluators to evaluate four popular MLLMs on the Chart w/o Code setting with criteria denoted in \Cref{fig:human_eval}. The results are listed below \Cref{tab:human_eval}, which show alignments between code-level with the human evaluations.

\begin{table}[h] % htbp 是浮动体位置参数 (here, top, bottom, page)
  \centering % 使表格居中
  \begin{tabular}{lccc}
    \toprule % booktabs 提供的顶部线条
    \textbf{Model} & {\textbf{Editing}} & {\textbf{Other}} & {\textbf{Sum}} \\ % 表头，S 列中的 \textbf 需要额外花括号
    \midrule % booktabs 提供的中间线条
    GPT-4o          & 3.88 & 3.07 & 6.95 \\
    InternVL2.5-72B & 3.08 & 2.36 & 5.44 \\
    InternVL2.5-8B  & 1.62 & 1.25 & 2.87 \\
    ChartMoE        & 1.21 & 1.03 & 2.24 \\
    \bottomrule % booktabs 提供的底部线条
  \end{tabular}
  \caption{The results of human evaluation on four popular MLLMs} % 表格标题
  \label{tab:human_eval}
  \vspace{-10pt}
\end{table}

\subsection{Prompts}
For the chart editing task, we utilize prompts in \Cref{fig:code_generation_prompt} to instruct MLLMs to generate the edited code.
For evaluation metrics, we utilize prompts in \Cref{fig:code_level_prompt} and \Cref{fig:chart_level_prompt} to evaluate the code-level and chart-level scores. 
% Figure 10 shows the prompt used for evaluating code generation, while Figure 11 focuses on the prompt used for code-level evaluation. Finally, Figure 12 illustrates the prompt for chart-level evaluation, which helps assess the quality and relevance of the generated charts.

\begin{figure}[h] % htbp 是图片位置建议：h (here), t (top), b (bottom), p (page of floats)
    \begin{tcolorbox}[
        colback=white,          % 背景颜色设置为白色
        colframe=black,         % 边框颜色设置为黑色
        title=Human Evaluation Instruction % tcolorbox 的标题
    ]

    You are a human evaluator. Please evaluate and score the completion degree of the chart editing based on the following scoring criteria.

    \medskip % 在第一个问题前添加一些垂直间距
    \noindent\textbf{1. How well does the edited chart follow the given instructions?}
    \medskip % 在表格前添加一些垂直间距

    % 第一个评分表格
    \noindent % 防止表格缩进
    \begin{tabularx}{\linewidth}{cX} % 表格宽度为当前行宽，c列居中，X列自动填充并换行
    \hline
    \textbf{Score} & \textbf{Criteria} \\
    \hline
    \textbf{4}     & The edited instructions are perfectly performed with no errors. \\
    \hline
    \textbf{4} & Instructions are generally followed, with minor deviations (e.g., small formatting differences). \\
    \hline
    \textbf{3} & Key edits are applied, but some instructions are missed or incorrect. \\
    \hline
    \textbf{2} & Only a few edits are correct; major deviations remain. \\
    \hline
    \textbf{1} & Minimal adherence; the generated chart barely reflects requested changes. \\
    \hline
    \textbf{0} & Exec failure or nothing corresponding to the instructions are edited. \\
    \hline
    \end{tabularx}
    \medskip

    \noindent\textbf{2. How well does the edited chart retain non-instructed editing elements from the original?}
    \medskip

    % 第二个评分表格
    \noindent % 防止表格缩进
    \begin{tabularx}{\linewidth}{cX}
    \hline
    \textbf{Score} & \textbf{Criteria} \\
    \hline
    \textbf{5} & All the elements (labels, scales, styles) except those instructed for editing remain identical. \\ % 修正了 "eidting" -> "editing"
    \hline
    \textbf{4} & Minor unintended changes (e.g., slight font size differences). \\
    \hline
    \textbf{3} & Some unmodified elements are altered but the charts are generally similar. \\
    \hline
    \textbf{2} & Significant unintended changes (e.g., missing lots of details). \\
    \hline
    \textbf{1} & Most original elements are lost or distorted. \\
    \hline
    \textbf{0} & Exec failure or all the other components are failed to construct. \\
    \hline
    \end{tabularx}

    \medskip
    \noindent\textbf{Final Score:} Sum of both dimensions (max 10 points).

    \end{tcolorbox}
    \vspace{-5pt}
    \caption{Human evaluator evaluation instructions} % 图的标题
    \label{fig:human_eval}               % 用于交叉引用的标签
\end{figure}

\begin{figure*}[t]
\begin{tcolorbox}[colback=white, colframe=black, title=Prompt for Edited Code Generation]

\textbf{Edited code generation without code (Chart w/o Code)}\\
You are an expert Python developer specializing in generating matplotlib code based on style modification instructions.   I will provide you with a reference image and a set of style modification instructions.  Your task is to generate the corresponding Python code according to the modification instructions and ensure that other parts remain unchanged except for the modified content.  The required modifications are as follows: \{\textcolor{blue}{instruction}\} and figure size is set to \{\textcolor{blue}{figsize}\}, and the generated code should be executable without requiring further modifications. Now, generate the Python code that produces a chart reflecting these changes. The code should be wrapped in \verb|```python\n```| \\

\textbf{Edited code generation with code (Chart w/ Code)}\\
You are an expert Python developer specializing in generating matplotlib code based on style modification instructions.   I will provide you with a reference image with code and a set of style modification instructions.  Your task is to generate the corresponding Python code according to the modification instructions and ensure that other parts remain unchanged except for the modified content.  The reference is: \{\textcolor{blue}{code}\} and the required modifications are as follows: \{\textcolor{blue}{instruction}\} and figure size is set to \{\textcolor{blue}{figsize}\}, and the generated code should be fully executable without requiring further modifications. Now, generate the Python code that produces a chart reflecting these changes. The code should be wrapped in \verb|```python\n```| \\

\textbf{Edited code generation without code of CoT (Chart w/o Code with CoT)}\\
You are an expert Python developer specializing in generating matplotlib code based on style modification instructions. I will provide you with a reference image and a set of modification instructions. Your task is to generate the corresponding Python code according to the modification instructions and ensure that other parts remain unchanged except for the modified content. The required modifications are as follows: \{\textcolor{blue}{instruction}\} and figure size is set to \{\textcolor{blue}{figsize}\}, and the generated code should be fully executable without requiring further modifications.
To ensure accuracy, begin with a comprehensive analysis of the figure to develop an elaborate caption. This caption should cover, but not be limited to, the following aspects:\\
	1.	Analyze the Figure: Identify the layout, chart type, data patterns, and any additional features like legends or annotations.\\
	2.	Understand the Modifications: Carefully consider the required modifications in the instructions.\\
	3.	Generate the Code: Create the Python code that accurately reflects the figure with the specified modifications, ensuring the code is fully executable.\\
Once you've completed these steps, generate the corresponding Python code. The code should be wrapped in \verb|```python\n```| 

\end{tcolorbox}
\caption{Prompt for generating the edited code.}
\label{fig:code_generation_prompt}
\end{figure*}

\begin{figure*}[t]
\begin{tcolorbox}[colback=white, colframe=black, title=Prompt for Code Level Evaluation]
You are an expert evaluator tasked with assessing the performance of a model on a Python code generation task. 
You will be provided with the original Python code, the instructions given to the model, and the code generated by the model.\\
The original code: \{\textcolor{blue}{source code}\}\\
Instructions: \{\textcolor{blue}{description}\}\\
The generated code: \{\textcolor{blue}{generated code}\}\\

\textbf{Scoring Methodology:}\\
The AI-generated code score is based on the following criteria, totaling a score out of 100:\\
1. Modification Accuracy (50 points): \\
-- Does the model make accurate and comprehensive modifications based on the instructions?\\
2. Code Completeness (50 points): \\
-- Is the generated code completely detailed and precise?\\

\textbf{Evaluation:}\\
Compare the two Python code files and provide a detailed assessment. Use the following format for your response:
\\
Comments:\\
-- Modification Accuracy: your comment and subscore\\
-- Code Completeness: your comment and subscore\\
Score:\\
-- Your final score out of 100\\

Please ensure the evaluation is clear and comprehensive.

\end{tcolorbox}
\caption{Prompt for code-level evaluation. \textcolor{blue}{source code} and \textcolor{blue}{generated code} are the human annotated and MLLM generated code receptively. \textcolor{blue}{description} is the editing instructions.}
\label{fig:code_level_prompt}
\end{figure*}

\begin{figure*}[t]
\begin{tcolorbox}[colback=white, colframe=black, title=Prompt for Chart Level Evaluation]
You are an excellent judge at evaluating visualization chart plots. The first image (reference
image) is created using ground truth matplotlib code, and the second image (AI-generated
image) is created using matplotlib code generated by an AI assistant. Your task is to score
how well the AI-generated plot matches the ground truth plot.\\
 \textbf{Scoring Methodology:}\\
The AI-generated image’s score is based on the following criteria, totaling a score out of 100
points:\\
1. Chart Types (20 points): \\
-- Does the AI-generated image include all chart types present in
the reference image (e.g., line charts, bar charts, etc.)?\\
2. Layout (10 points): \\
-- Does the arrangement of subplots in the AI-generated image match the
reference image (e.g., number of rows and columns)?\\
3. Text Content (20 points): \\
-- Does the AI-generated image include all text from the reference
image (e.g., titles, annotations, axis labels), excluding axis tick labels?\\
4. Data (20 points): \\
-- How accurately do the data trends in the AI-generated image resemble
those in the original image and is the number of data groups the same as in the reference
image?\\
5. Style (20 points):\\
-- Does the AI-generated image match the original in terms of colors (line
colors, fill colors, etc.), marker types (point shapes, line styles, etc.), legends, grids, and other
stylistic details?\\
6. Clarity (10 points):\\
-- Is the AI-generated image clear and free of overlapping elements?\\
\textbf{Evaluation:}\\
Compare the two images head to head and provide a detailed assessment. Use the following
format for your response:
\\
Comments:\\
-- Chart Types: your comment and subscore\\
-- Layout: your comment and subscore\\
-- Text Content: your comment and subscore\\
-- Data: your comment and subscore\\
-- Style: your comment and subscore\\
-- Clarity: your comment and subscore\\
Score: \\
-- Your final score out of 100\\

Please use the above format to ensure the evaluation is clear and comprehensive.

\end{tcolorbox}
\caption{Prompt for chart-level evaluation.}
\label{fig:chart_level_prompt}
\end{figure*}
\end{document}